\documentclass{article}

\PassOptionsToPackage{table}{xcolor}

\usepackage{microtype}
\usepackage{graphicx}
\usepackage{subcaption}
\usepackage{booktabs} 

\usepackage{hyperref}

\usepackage[preprint]{icml2026}

\usepackage{amsmath}
\usepackage{amssymb}
\usepackage{mathtools}
\usepackage{amsthm}
\usepackage{booktabs}
\usepackage{multirow}
\usepackage{xspace}

\usepackage[capitalize,noabbrev]{cleveref}

\theoremstyle{plain}

\theoremstyle{definition}

\theoremstyle{remark}

\usepackage[disable,textsize=tiny]{todonotes}

\icmltitlerunning{DynaTok: Token-Based 4D Reconstruction from Partial Point Clouds}

\newcommand{\oursName}{DynaTok\@\xspace}

\begin{document}

\twocolumn[
  \icmltitle{DynaTok: Token-Based 4D Reconstruction from Partial Point Clouds}



  \icmlsetsymbol{equal}{*}
  
  \begin{icmlauthorlist}
    \icmlauthor{Weirong Chen}{tum,google,mcml}
    \icmlauthor{Keisuke Tateno}{google}
    \icmlauthor{Hidenobu Matsuki}{google}
    \icmlauthor{Michael Niemeyer}{google}
    \icmlauthor{Daniel Cremers}{tum,mcml}
    \icmlauthor{Federico Tombari}{tum,google,mcml}
  \end{icmlauthorlist}

  \icmlaffiliation{tum}{TU Munich}
  \icmlaffiliation{google}{Google}
  \icmlaffiliation{mcml}{Munich Center for Machine Learning}
  \icmlcorrespondingauthor{Weirong Chen}{weirong.chen@tum.de}


  \vskip 0.3in
]



\printAffiliationsAndNotice{}  

\begin{abstract}

We address 4D reconstruction from partial point cloud sequences, where depth-sensor observations are incomplete, unordered, and lack explicit temporal correspondences. This geometry-only setting is challenging due to missing observations and ambiguous dynamics. While recent progress has largely relied on image-based methods, existing point-based approaches typically focus on single objects, assume relatively complete inputs, or require explicit correspondences.
To address these limitations, we propose DynaTok, a point-based framework for correspondence-free 4D reconstruction from partial point cloud sequences without images. DynaTok encodes frames into compact latent tokens, aggregates incomplete observations over time with a Transformer-based spatiotemporal encoder, and decouples geometry and motion through residual tokens in a unified model. A flow-matching decoder then reconstructs complete, temporally consistent 4D point-cloud sequences conditioned on the latent tokens.
Experiments on object- and scene-level benchmarks demonstrate improved reconstruction quality and temporal coherence from partial point cloud observations. Project page: https://wrchen530.github.io/dynatok/.
\end{abstract}    
\section{Introduction}
\label{sec:intro}

\begin{figure}[t]
\centering
\includegraphics[width=\linewidth]{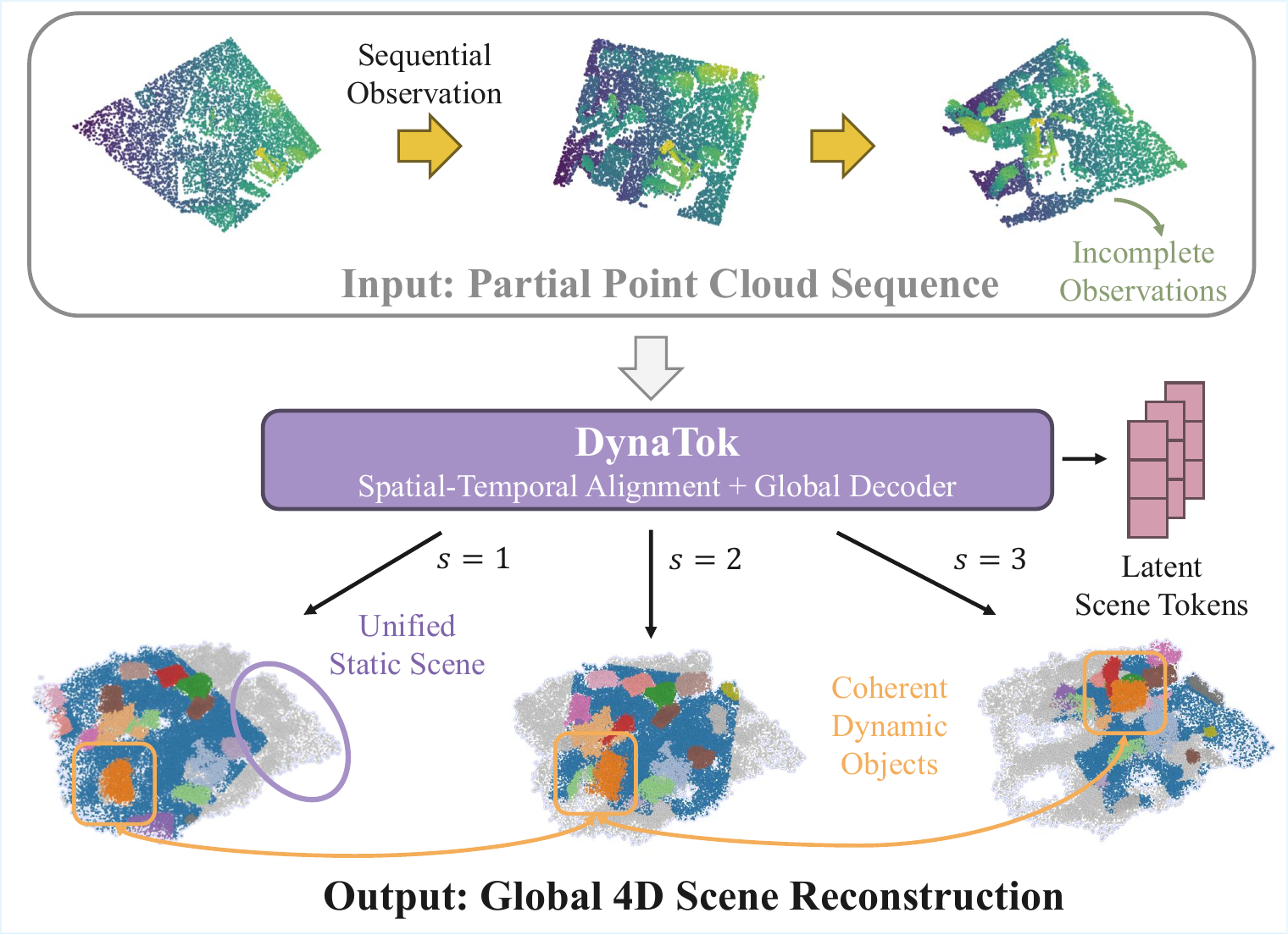}
\caption{
\textbf{DynaTok} reconstructs coherent 4D scenes from partial, unordered, and correspondence-free point cloud sequences. It temporally aggregates incomplete observations via spatiotemporal alignment and a global decoder, enabling consistent recovery of static background structure and dynamic objects even when large regions are unobserved.
For visualization, we show the global 4D scene reconstruction at each time step: the colored region corresponds to the area observed in the partial input at time $s$, while the white regions are recovered through the temporal fusion.\textbf{ Colors and segmentation labels are only used for visualization and are not provided to the model.}
}%
\label{fig:teaser}
\vspace{-0.4em}
\end{figure}

Reconstructing dynamic 3D scenes over time, which is commonly referred to as 4D reconstruction, is a fundamental problem in computer vision with broad applications in robotics~\cite{rajivc2025multi, huang2026pointworld}, augmented reality~\cite{wang2025shape}, and scene understanding~\cite{zhang2025egogaussian,zhou2024drivinggaussian}. Recent advances have been largely driven by methods that operate on dense visual observations, such as images or videos~\cite{zhang2024monst3r, feng2025st4rtrack,li2025megasam,ren2024l4gm,wu2025cat4d}, enabling high-fidelity reconstruction of dynamic geometry under controlled settings. However, these approaches typically assume visual input with rich texture and appearance cues. 
In contrast, point clouds are the native output of depth sensors and provide a scalable, geometry-centric representation without relying on appearance cues or surface connectivity. Despite their suitability for real-world sensing, existing point-based learning methods are largely restricted to static scenes or object-level dynamics and struggle to scale to complex, open-world dynamic environments.

In many practical scenarios, especially those involving depth sensors or multi-view geometry pipelines, the available observations at test time take the form of \textit{partial point clouds}. These point clouds are incomplete due to limited sensor coverage, occlusions, and viewpoint changes, are unordered, and lack explicit point-to-point correspondence across time. Under this setting, 4D reconstruction faces several fundamental challenges.
First, individual frames may entirely miss large regions of the scene, making per-frame reconstruction fundamentally insufficient, particularly for dynamic objects.
Second, without image appearance or tracking cues, it is inherently ambiguous to distinguish static scene structure from dynamic objects, as missing points may result from occlusion, motion, or viewpoint changes.
Third, since the identity of points is not preserved across frames, temporal correspondence cannot be assumed, and naive temporal fusion often leads to inconsistent geometry and motion. Together, these challenges make coherent 4D reconstruction from partial point cloud sequences fundamentally difficult.

In this work, we study 4D reconstruction from partial, correspondence-free point cloud inputs, with a particular focus on how temporal aggregation across frames can compensate for missing geometry. Rather than assuming that each time step provides a complete or dense observation, our goal is to recover a temporally consistent 4D representation even when objects or regions are completely unobserved in individual frames (see~\Cref{fig:teaser}). This setting poses unique challenges that are not adequately addressed by existing methods, which often rely on complete inputs or focus on single-object dynamics~\cite{cao2024motion2vecsets,niemeyer2019occupancyflow,tang2021lpdc,lei2022cadex}. Moreover, many of these methods require watertight mesh supervision, which limits their applicability to real-world scanned data.

To address this challenge, we propose \oursName, a token-based framework for temporally aggregating partial point cloud observations. The key idea is to encode sparse and irregular point cloud inputs into a compact set of temporally aligned latent tokens shared across frames. By aggregating complementary information over time, these tokens enable robust 4D reconstruction from incomplete observations.
We further decouple the latent space into geometry and motion tokens, enabling disentangled modeling of shape and dynamics. Unlike prior approaches that rely on two separate networks for shape reconstruction and deformation modeling~\cite{cao2024motion2vecsets,jiang2026mesh4d,zhang2025gvfd}, our residual token formulation achieves explicit geometry–motion decoupling within a single unified model.
Conditioned on the aggregated latent tokens, \oursName reconstructs coherent 4D scene representations using a point flow-matching decoder~\cite{lipmanflow}, which produces temporally consistent point clouds and can be trained using only point cloud supervision. This design avoids reliance on watertight meshes or explicit correspondences and provides a unified framework for both object-level and scene-level 4D reconstruction.

In summary, our contributions are threefold. 
(i) We formulate the practical yet underexplored problem of global 4D
reconstruction from partial, unordered, and correspondence-free point cloud
sequences.
(ii) We propose a unified token-based 4D representation and pipeline that
enables correspondence-free temporal aggregation and geometry--motion
decomposition under point-cloud-only supervision, without requiring watertight
meshes or canonical templates.
(iii) We validate the framework on both object- and scene-level dynamic
benchmarks through qualitative and quantitative evaluations.

\section{Related Work}
\label{sec:related_work}

\subsection{3D Generation and Latent Representations}
3D generative models aim to learn compact latent representations of shapes and scenes. Early work employs VAEs and GANs to model distributions over 3D geometry~\cite{wu2016learning,zhu2018visual,pavllo2020convolutional,kosiorek2021nerf,li2021sp,chan2022efficient,gao2022get3d}, while recent advances leverage diffusion models to generate high-fidelity 3D content. One line of work combines diffusion with neural rendering, generating multi-view images that are reconstructed into 3D via differentiable rendering~\cite{liu2023zero,gao2024cat3d,szymanowicz2025bolt3d,tang2024lgm,hong2024lrm,voleti2024sv3d,chen2024mvsplat360}. 

Another direction directly generates 3D representations such as point clouds or meshes using diffusion~\cite{zhang20233dshape2vecset,li2025triposg,jun2023shap,ren2024xcube,vahdat2022lion,wu2024direct3d,zhang2024clay,chang20243d,zheng2023locally,gupta20233dgen,xiang2025structured}. 
Closely related, NOVA3R~\cite{chennova3r} learns a global latent representation for static 3D scenes in a two-stage framework and supports image-conditioned reconstruction. 
Despite these advances, most 3D generative models are designed for static
shapes or bounded scenes, with limited support for temporally evolving 4D
geometry from partial, correspondence-free inputs. In contrast, we study this latent-representation perspective in the dynamic 4D setting, where the input is a partial, correspondence-free point-cloud sequence.

\subsection{4D Reconstruction from Video}
Reconstructing 4D scenes from video aims to recover temporally coherent 3D geometry
and motion from visual observations, providing a dynamic representation of objects or scenes over time~\cite{wang2025shape,li2025megasam,zhang2024monst3r,feng2025st4rtrack,chen2025back,zhang2025efficiently,qian2026flow4r,karhade2025any4d}.
Recent advances in image-based 4D reconstruction are driven by generative methods that model dynamic geometry from multi-view video, typically focusing on single objects, explicit surface representations (e.g., meshes or canonical templates), and dense visual supervision to decouple shape and motion. Early works incorporate generative or physical priors to regularize motion during optimization~\cite{jiangconsistent4d,ren2023dreamgaussian4d,sang2025twosquared}, while more recent approaches leverage diffusion models to directly recover time-varying geometry~\cite{ren2024l4gm,liang2024diffusion4d,pan2024efficient4d,xiesv20254d,zhang20244diffusion,jiang2026mesh4d,zhang2025gvfd}. However, these methods largely assume image inputs, single-object scenes, or explicit correspondences at inference, limiting their applicability to our setting of partial, correspondence-free point clouds.

Beyond 4D object generation, scene-level 4D reconstruction from images is commonly formulated as recovering dense point maps~\cite{Kopf_2021_CVPR,zhang2022structure,zhang2024monst3r,li2025megasam,wimbauer2025anycam} or Gaussian Splats~\cite{wu20244d,wang2025shape,yang2024real,matsuki20254dtam} per time step. These methods exploit multi-view consistency, optical flow, and appearance cues to recover dynamic scenes, with recent work further incorporating dense correspondence and scene flow for more complete 4D reconstruction~\cite{sucar2025dynamic,wang2025c4d,karhade2025any4d,feng2025st4rtrack,zhang2025efficiently}. By relying on dense image observations and rich appearance cues, these methods address a different input regime from the geometry-only point-cloud setting considered here.

\subsection{4D Reconstruction from Point Clouds}
In contrast to image-based methods, recent work has explored 4D reconstruction from point cloud observations to address occlusions, limited coverage, and viewpoint changes in real-world systems~\cite{zheng2023rcfusion}. Such point clouds may be obtained from depth sensors~\cite{ranftl2020towards,yang2024depth,yin2023metric3d}, LiDAR~\cite{huang2023neuralcvpr,huang2023neural}, or sparse geometric cues such as multi-view anchor points~\cite{dong2021shape} and motion capture systems~\cite{alexiadis2016integrated}. The core challenge is to aggregate incomplete geometry over time to recover coherent 4D structure. Classical fusion and SLAM-based methods show that temporal aggregation substantially improves static reconstruction under sparse observations~\cite{izadi2011kinectfusion,dai2017bundlefusion}, and recent learning-based approaches extend these principles to dynamic point cloud sequences.

A common approach to 4D reconstruction disentangles static structure from dynamic deformation, often via canonical-to-deformed formulations such as Occupancy Flow~\cite{niemeyer2019occupancyflow} and related correspondence- or deformation-based object-level methods~\cite{tang2021lpdc,lei2022cadex,cao2024motion2vecsets,wang2025canfields}. 
Among these, Motion2VecSets~\cite{cao2024motion2vecsets} is most closely
related to our work, but it infers a canonical shape from only the first frame
and generates later frames as deformations of this fixed reference, whereas our
one-stage alignment module jointly aggregates geometry across frames to better
handle incomplete first-frame observations and correspondence-free partial
point-cloud sequences.
Scene-level approaches adopt explicit grid or implicit representations for scalability~\cite{Huang_2025_CVPR,zhong20243d}, but usually rely on relatively complete inputs such as LiDAR sweeps. In contrast, we address 4D reconstruction from partial, correspondence-free point clouds derived from depth maps with severe occlusions and viewpoint changes.

\section{Method}
\label{sec:method}

\begin{figure*}[t]
\centering
\includegraphics[width=\linewidth]{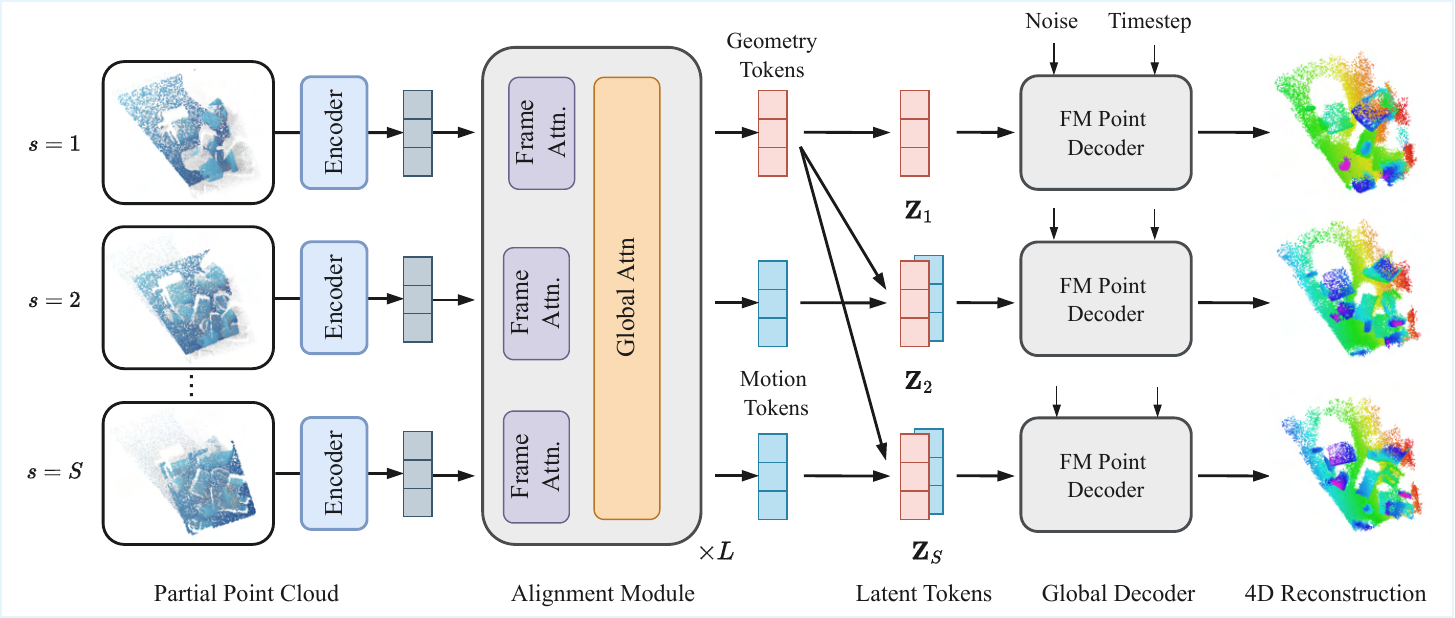}
\caption{\textbf{Overview of the DynaTok Pipeline.} Given a sequence of incomplete point clouds, we extract per-frame point tokens and process them with a spatiotemporal alignment module to integrate temporal information. Using a residual token design, we obtain geometry tokens defining the canonical space at the reference frame ($s=1$) and motion tokens for subsequent frames ($s>1$), enabling joint shape and motion modeling without introducing an additional network. The resulting latent token $\mathbf{Z}_s$ encodes 4D information and is fed into the flow-matching decoder to recover the global geometry at time step $s$. All components are trained from scratch.
}%
\label{fig:pipeline}
\vspace{-0.4em}
\end{figure*}

Our goal is to reconstruct a coherent 4D scene representation from a sequence of partial, unordered, and correspondence-free point clouds. The main challenge lies in temporal aggregation: integrating incomplete and inconsistent observations across time to recover missing geometry and coherent motion in dynamic scenes. We address this challenge by proposing a 4D point tokenizer that embeds input point clouds into a compact latent space, enabling the recovery of global scene geometry at each time step. In the following, we first formalize the problem setting (\Cref{sec:method_problem}), then describe the latent temporal aggregation mechanism (\Cref{sec:method_encoder}) and the conditional reconstruction process (\Cref{sec:method_decoder}). An overview of the pipeline is shown in \Cref{fig:pipeline}.

\subsection{Problem Formulation}
\label{sec:method_problem}

We consider the task of reconstructing a complete and temporally consistent 4D scene from a sequence of partial point cloud observations. 
The input is a sequence of dynamic point clouds $\mathcal{X}= \{\mathbf{X}_s\}_{s=1}^S$, where $\mathbf{X}_s \in \mathbb{R}^{N_{\text{in}} \times 3}$ denotes the 3D points observed at time step $s$. 
Each $\mathbf{X}_s$ can be partial, unordered, and correspondence-free, reflecting realistic observations from depth sensors or multi-view reconstruction pipelines.
The target is to recover a temporally consistent and complete 4D scene representation $\mathcal{Y} = \{\mathbf{Y}_s\}_{s=1}^S, \mathbf{Y}_s \in \mathbb{R}^{N_{\text{out}} \times 3}$, where each $\mathbf{Y}_s$ represents the complete scene geometry at time step $s$. Importantly, the reconstruction at each time step must be consistent with the underlying scene motion, even when the corresponding geometry is unobserved in the input.

\begin{figure}[t]
\centering
\includegraphics[width=\linewidth]{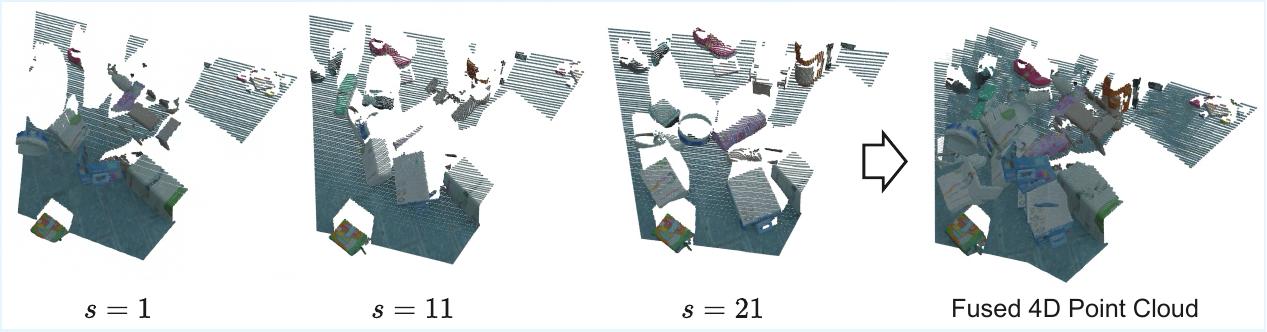}\vspace{-.3cm}
\caption{\textbf{Illustration of 4D Fusion and Completion Task.} Given the partial observation at each time step, the goal is to model the global dynamic scene.}%
\label{fig:4dcomplete}
\vspace{-0.4em}
\end{figure}

This problem setting is substantially more challenging than per-frame completion~\cite{yu2021pointr, huang2020pf} or 4D autoencoders with complete geometry~\cite{zhang2025gvfd,wang2024occsora}. Successful reconstruction requires the model to (i) aggregate incomplete and inconsistent observations across time, (ii) implicitly distinguish static scene structure from dynamic objects without correspondence or tracking cues, and (iii) maintain temporal coherence of the reconstructed geometry across frames. These challenges cannot be addressed by naive temporal fusion or frame-wise reconstruction, and instead demand structured mechanisms for effective temporal aggregation under partial observations (see~\Cref{fig:4dcomplete}).

\subsection{Latent Temporal Aggregation}
\label{sec:method_encoder}

\paragraph{Per-Frame Token Extraction.}

To enable temporal aggregation over partial and correspondence-free point clouds, we first convert each input point cloud into a compact and structured latent representation. Given an input point cloud $\textbf{X}_s \in \mathbb{R}^{N_{\text{in}} \times 3}$ at time step $s$, we extract a fixed number of latent tokens that represent the geometric content of the frame in a permutation-invariant manner. 

Specifically, we select $M$ point queries as anchors using Farthest Point Sampling (FPS). Inspired by recent geometry autoencoders~\cite{li2025triposg,chennova3r}, these queries attend to the full input point cloud through a cross-attention layer, where the raw point positions serve as keys and values. Prior to attention, 3D point coordinates are embedded into features using Fourier features to enhance spatial expressiveness~\cite{li2025triposg}. The resulting position features are further refined using self-attention layers to capture local geometric context. The output is a set of per-frame latent tokens $\mathbf{F}_s \in \mathbb{R}^{M \times D_{\mathcal{Z}}}$, providing a compact and regularized representation of each partial point cloud. This tokenization step reduces the variability of raw point cloud inputs while preserving essential geometric structure, enabling efficient and consistent processing across time.

\paragraph{Spatial-Temporal Alignment.}

Given the per-frame tokens $\{\mathbf{F}_{s} \}_{s=1}^S$, we perform temporal aggregation using a transformer-based architecture~\cite{wang2025vggt} that jointly processes tokens across time. The goal is to integrate partial observations from different frames and aggregate geometric information into a coherent latent representation. 

Each transformer layer alternates between per-frame self-attention and cross-frame global attention. Per-frame attention preserves spatial coherence within a frame, and the global attention allows tokens to propagate information across time in the latent space. Through this interleaved attention mechanism, geometric evidence observed in one frame can influence the latent representation of other frames, enabling recovery of missing regions under partial observations. 

Importantly, this aggregation operates entirely at the token level and does not assume explicit point-to-point correspondence across frames. Temporal consistency emerges implicitly through shared attention and repeated exposure to complementary observations. To encode spatial relationships, we extend rotary positional embeddings~\cite{su2024roformer} to 3D using the token query positions, allowing the model to reason about relative geometry during aggregation.

\paragraph{Latent Residual Decomposition.}

To model temporal variation in the aggregated latent representations, we adopt a simple residual formulation across time. We treat the latent tokens corresponding to the first frame ($s=1$) as a reference representation, which serves as a canonical anchor for the sequence. This reference does not introduce a separate latent space or specialized tokens; all frames are represented in the same latent space.

Formally, we represent the latent tokens at time $s$ as: 
\begin{equation}
\mathbf{Z}_s = \mathbf{G} + \mathbf{H}_s,
\end{equation}
where $\mathbf{G}$ denotes the reference latent representation and $\mathbf{H}_s$ captures the residual changes at time $s$. For the reference frame, the residual is set to zero, i.e., $\mathbf{H}_1 = 0$.
This formulation encourages persistent structure to be shared across time while allowing flexible temporal variation through residual offsets.
The resulting latent representations $\{\mathbf{Z}_s\}$ compactly encode both persistent scene structure and time-varying geometry, and are used by the decoder for conditional 4D reconstruction.

Importantly, the proposed residual formulation introduces an explicit yet simple separation between time-invariant structure and time-varying components within a shared latent space. This separation is realized through the reference–residual parameterization, without introducing additional tokens or auxiliary objectives. The model is trained end-to-end using only supervision from the global point cloud at each time step, under which it automatically learns to allocate persistent geometry to the reference component and temporal variation to the residuals, without requiring motion labels.

\subsection{Conditional Global Decoder}
\label{sec:method_decoder}

Given the temporally aggregated latent representation $\mathbf{Z}_s$ at time step $s$, we introduce a global 3D decoder to reconstruct the complete scene geometry at time step $s$ as a point cloud $\mathbf{Y}_s \in \mathbb{R}^{N_{\text{out}}\times 3}$. We formulate this global reconstruction as learning a conditional distribution over point sets $p(\mathbf{Y}_s | \mathbf{Z}_s)$, where the latent tokens $\mathbf{Z}_s$ explicitly encode the canonical scene geometry via $\mathbf{G}$ and the time-varying deformation via the residual motion tokens $\mathbf{H}_s$. This formulation supports globally consistent reconstruction over unordered point sets with variable cardinality, enabling the decoder to output point clouds at any density during inference.

To parameterize this conditional distribution, we adopt a conditional flow-matching decoder~\cite{chang20243d}, which models point cloud generation as a continuous transformation from a simple prior to the target scene geometry, conditioned on latent tokens $\mathbf{Z}_s$. This design is well aligned with our setting: flow matching naturally supports variable point-cloud cardinalities and avoids explicit point correspondences, making it well suited to correspondence-free reconstruction.

\paragraph{Conditional Flow Matching.}
We denote $s$ as the discrete frame index corresponding to the input sequence, and $t \in [0, 1]$ as the continuous time variable used in the flow-matching process~\cite{lipmanflow}. Given a target point cloud $\mathbf{x}_1$ and a noise sample $\boldsymbol{\epsilon}$, we define a linear interpolation path from noise to data as
\begin{equation}
\mathbf{x}_t = (1-t)\boldsymbol{\epsilon} + t\mathbf{x}_1 .
\label{eq:fm_path}
\end{equation}
Thus, $t=0$ corresponds to the prior distribution and $t=1$ corresponds to the target point-cloud distribution. The decoder learns a conditional velocity field $\Phi_{\text{dec}}(\mathbf{x}_t, t, \mathbf{Z}_s)$, which defines the ordinary differential equation (ODE)
\begin{equation}
\frac{d \mathbf{x}_t}{dt} = \Phi_{\text{dec}}(\mathbf{x}_t, t, \mathbf{Z}_s).
\end{equation}
For this linear path, the target velocity is given by
$\mathbf{v}_{\text{target}} = \mathbf{x}_1 - \boldsymbol{\epsilon}$.
The decoder is trained to predict this velocity field conditioned on the scene token $\mathbf{Z}_s$, thereby transporting samples from the prior distribution to the target point-cloud distribution.

\paragraph{Training Objective.}

We train our decoder using the flow-matching loss as the sole loss term, without requiring any additional supervision. For each time step $s$, we randomly sample a target point cloud $\mathbf{x}_1 \sim \mathbf{Y}_s$, a noise sample $\boldsymbol{\epsilon} \sim \mathcal{U}([-1,1]^3)$ from a uniform cube prior, and a flow time $t \sim \mathcal{U}(0,1)$. 
Given the interpolation path in~\cref{eq:fm_path}, the training objective minimizes the distance between the predicted and target velocity fields:
\begin{equation}
\mathcal{L}_{\mathrm{FM}} =
\mathbb{E}_{t, \mathbf{x}_1, \boldsymbol{\epsilon}, \mathbf{Z}_s}
\left[
\left\|
\Phi_{\mathrm{dec}}(\mathbf{x}_t, t, \mathbf{Z}_s)
-
(\mathbf{x}_1 - \boldsymbol{\epsilon})
\right\|_2^2
\right].
\end{equation}
This objective encourages the decoder to reconstruct complete and temporally consistent global geometry conditioned on the aggregated scene representation, without additional supervision.
During inference, we sample $\boldsymbol{\epsilon} \sim \mathcal{U}([-1,1]^3)$ and integrate the learned ODE forward from $t=0$ to $t=1$, starting from $\mathbf{x}_{t=0}=\boldsymbol{\epsilon}$. This yields the reconstructed point cloud $\mathbf{Y}_s$.

\paragraph{Implementation Details.}
All models are implemented in PyTorch and trained on 8 NVIDIA H100 GPUs. We use the AdamW optimizer with a learning rate of $10^{-3}$ and a linear warm-up schedule. Models are trained for 250k iterations with a batch size of 8 sequences per GPU. Input and target point clouds are normalized using median-based scaling. During training, we randomly sample $S=8$ frames per sequence, while evaluation is performed on sequences of 16 frames. Unless otherwise specified, the input partial point cloud contains 8,192 points per frame. We use $M=512$ tokens per frame for latent representation. Additional implementation details are provided in the supplementary material. We plan to release the code upon acceptance.

\section{Experiments}
\label{sec:experiment}

\subsection{Experimental Setup}

We evaluate DynaTok on both scene-level and object-level dynamic point cloud benchmarks, focusing on the setting of partial, correspondence-free point cloud inputs. All methods are evaluated on their ability to reconstruct temporally coherent 4D geometry from incomplete observations.

\paragraph{Datasets.} 
For object-level reconstruction, we adopt \textbf{DeformingThings4D–Animals (DT4D-A)}~\cite{li20214dcomplete}, a standard benchmark used by prior 4D point-based methods~\cite{cao2024motion2vecsets}. The dataset contains 1,972 sequences of deformable humanoids and animals exhibiting articulated and non-rigid motion. Each sequence consists of a single object without background, allowing us to isolate deformation modeling without scene-level clutter. We follow prior work and use point clouds back-projected from synthetic RGB-D scans as input.

For scene-level reconstruction, we use the \textbf{Kubric} dataset~\cite{greff2021kubric}, which contains complex interactions of static and dynamic objects from Google Scanned Objects~\cite{downs2022google}. It provides ground-truth complete point clouds for both static and moving components, enabling evaluation of 4D fusion and completion under partial observations. Compared to dynamic video datasets such as DAVIS~\cite{perazzi2016benchmark}, Kubric presents more challenging scenarios with fast motion, multiple independently moving objects, and camera motion. We follow the MOVi-F setup, using 10K training sequences and 200 test sequences, each with 30 frames at 10 FPS captured by a moving camera.

\paragraph{Baselines.}
To our knowledge, no prior method directly addresses 4D reconstruction from partial, correspondence-free point clouds at both object and scene levels as in our setting. We therefore compare DynaTok with the closest baseline, Motion2VecSets~\cite{cao2024motion2vecsets}, which models temporal dynamics but is limited to single objects and fixed-size inputs (512 points per frame). Image-based 4D generative methods such as GVFD~\cite{zhang2025gvfd} require explicit correspondences and complete shapes and are thus inapplicable. We also include representative 3D point VAEs—3DShape2VecSet~\cite{zhang20233dshape2vecset}, TripoSG~\cite{li2025triposg}, and TRELLIS~\cite{xiang2025trellis}—applied per frame as reference baselines to quantify the effect of missing temporal aggregation.
We use pretrained models and recommended settings when available, as baseline methods typically require complete mesh supervision for training.

\paragraph{Evaluation Metrics.}
We evaluate reconstruction quality using standard geometric metrics, consistent with prior work such as CUT3R~\cite{wang2025continuous}. Specifically, we report Accuracy and Completeness, defined as the one-sided Chamfer Distance from prediction to ground truth and vice versa, respectively. We also report Normal Consistency, which measures the alignment of surface normals and reflects local geometric fidelity. For point cloud outputs, we estimate normals using Open3D~\cite{zhou2018open3d}.

\begin{table}[t]
\centering
\small
\caption{\textbf{Object-Level Reconstruction Results on DT4D.}
Our method improves over object-level 4D baselines.}
\label{tab:dt4d_depth_8192}
\setlength\tabcolsep{2pt}
\resizebox{\columnwidth}{!}{
\begin{tabular}{@{}lcccccc@{}}
\toprule
\multirow{2}{*}{Method}
& \multicolumn{3}{c}{Unseen Motion}
& \multicolumn{3}{c}{Unseen Individual} \\
\cmidrule(lr){2-4} \cmidrule(l){5-7}
& Acc. $\downarrow$ & Comp. $\downarrow$ & NC $\uparrow$
& Acc. $\downarrow$ & Comp. $\downarrow$ & NC $\uparrow$ \\
\midrule

\rowcolor{gray!15}
\multicolumn{7}{@{}l@{}}{\textit{3D methods}} \\
Shape2VecSet~\cite{zhang20233dshape2vecset} & \textbf{0.022} & 0.174 & 0.703 & 0.030 & 0.167 & 0.696 \\
TripoSG~\cite{li2025triposg}        & 0.051 & 0.165 & 0.740 & 0.048 & 0.143 & 0.732 \\
TRELLIS~\cite{xiang2025trellis}        & 0.026 & 0.190 & 0.813 & 0.049 & 0.194 & 0.787 \\

\midrule
\rowcolor{gray!15}
\multicolumn{7}{@{}l@{}}{\textit{4D methods}} \\
Motion2VecSets~\cite{cao2024motion2vecsets} & 0.055 & 0.060 & 0.856 & 0.061 & 0.065 & 0.824 \\
Ours           & 0.023 & \textbf{0.021} & \textbf{0.914}
               & \textbf{0.027} & \textbf{0.026} & \textbf{0.877} \\
\bottomrule
\end{tabular}
}
\end{table}

\begin{figure}[t]
\centering
\includegraphics[width=\linewidth]{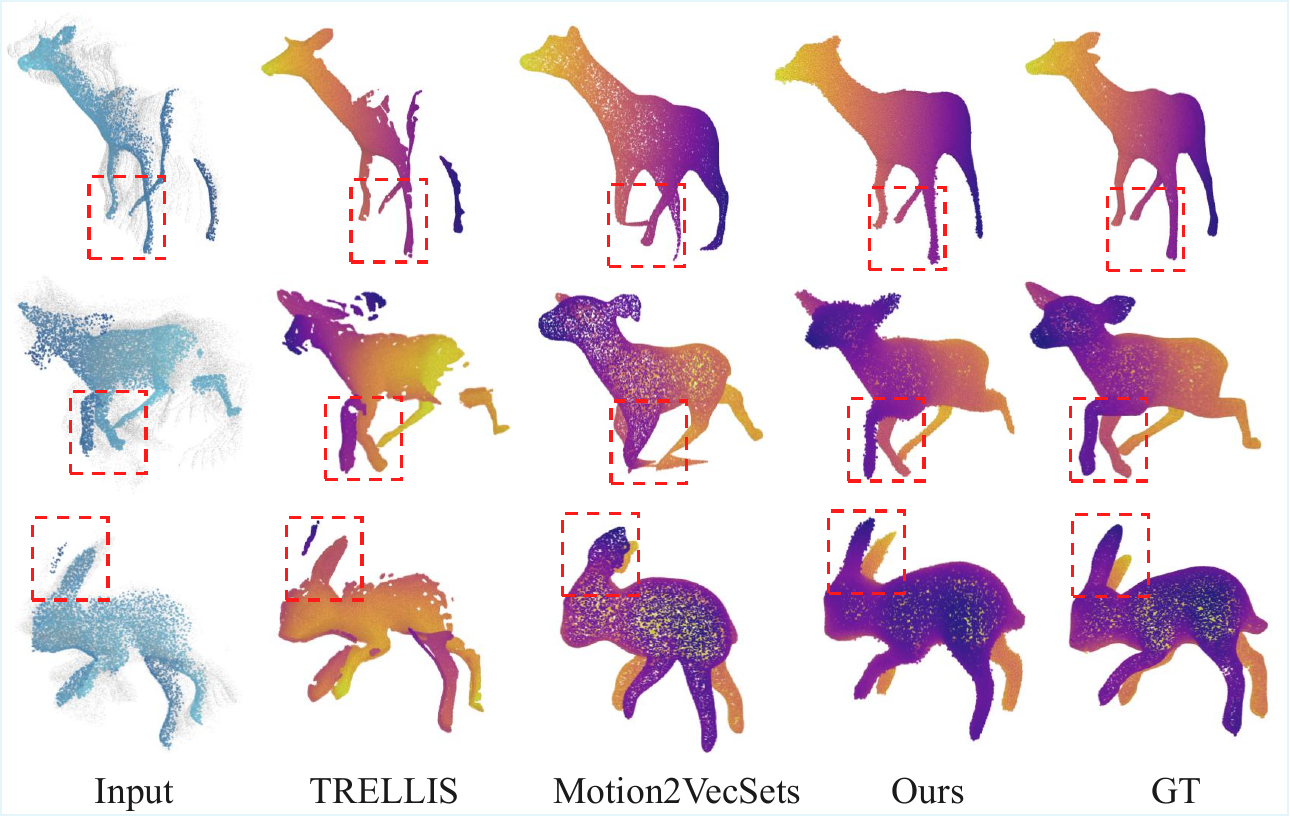}
\caption{\textbf{Qualitative Reconstruction Results on the DT4D Dataset~\cite{li20214dcomplete}}. We compare the partial point cloud input setting from depth maps. Our method achieves clearer geometry that matches the ground truth point locations. We include the partial inputs from other time steps in white for reference.} 
\label{fig:dt4d}
\vspace{-0.4em}
\end{figure}

\subsection{Object-level 4D Reconstruction}
We evaluate object-level 4D reconstruction on DT4D-A under the standard Unseen Motion and Unseen Individual settings~\cite{cao2024motion2vecsets,tang2021lpdc}. All methods are assessed using point-based metrics, with mesh outputs uniformly downsampled to $50k$ points for consistency. As shown in~\Cref{tab:dt4d_depth_8192}, DynaTok consistently outperforms prior 4D baselines, achieving much lower completeness error and higher normal consistency, highlighting its ability to recover missing geometry via temporal aggregation. 

\Cref{fig:dt4d} further shows qualitative results under depth-based partial input. Motion2VecSets~\cite{cao2024motion2vecsets} captures coarse motion but exhibits missing regions and inaccurate articulation, while DynaTok produces cleaner geometry and more coherent motion, closely matching ground truth despite severe occlusions. These results show that latent-space temporal aggregation enables robust 4D reconstruction from correspondence-free partial point clouds.

\begin{table}[t]
\centering
\small
\caption{\textbf{Scene-Level Reconstruction on Kubric.} Our method produces better reconstruction results than competing baselines.
Motion2VecSets is restricted to 512 input points and is therefore excluded from FG + BG.
}
\label{tab:kubric_depth_8192}
\setlength\tabcolsep{2pt}
\resizebox{\columnwidth}{!}{
\begin{tabular}{@{}lcccccc@{}}
\toprule
\multirow{2}{*}{Method}
& \multicolumn{3}{c}{Foreground}
& \multicolumn{3}{c}{Foreground + Background} \\
\cmidrule(lr){2-4} \cmidrule(l){5-7}
& Acc. $\downarrow$ & Comp. $\downarrow$ & NC $\uparrow$
& Acc. $\downarrow$ & Comp. $\downarrow$ & NC $\uparrow$ \\
\midrule

\rowcolor{gray!15}
\multicolumn{7}{@{}l@{}}{\textit{3D methods}} \\
Shape2VecSet~\cite{zhang20233dshape2vecset} & 0.008 & 0.027 & 0.680 & \textbf{0.009} & 0.026 & 0.724 \\
TripoSG~\cite{li2025triposg}        & 0.012 & 0.023 & 0.711 & 0.042 & 0.020 & 0.783 \\
TRELLIS~\cite{xiang2025trellis}        & 0.009 & 0.029 & 0.829 & 0.011 & 0.025 & 0.874 \\

\midrule
\rowcolor{gray!15}
\multicolumn{7}{@{}l@{}}{\textit{4D methods}} \\
Motion2VecSets~\cite{cao2024motion2vecsets} & 0.061 & 0.063 & 0.545 & -- & -- & -- \\
Ours           & \textbf{0.008} & \textbf{0.010} & \textbf{0.835}
               & 0.010 & \textbf{0.014} & \textbf{0.888} \\
\bottomrule
\end{tabular}
}
\end{table}

\begin{figure*}[t]
\centering
\includegraphics[width=\linewidth]{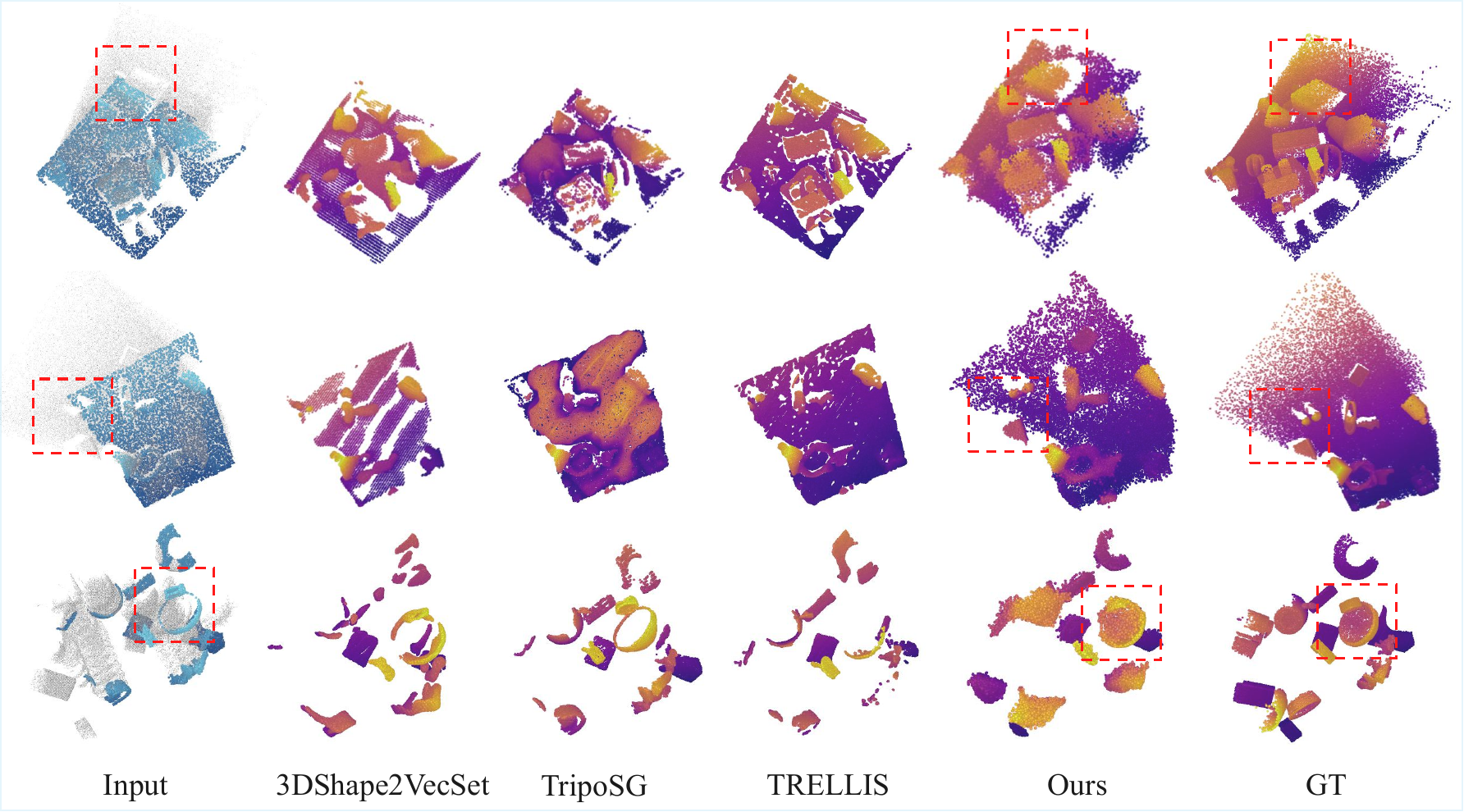}
\caption{\textbf{Qualitative Reconstruction Results on Kubric~\cite{greff2021kubric}}. The red boxes highlight regions where our method better preserves geometry than the baselines and effectively aggregates temporal information into a geometrically consistent canonical space. 
We include the partial inputs from other time steps in white for reference. }%
\label{fig:kubric_scene}
\vspace{-0.4em}
\end{figure*}

\subsection{Scene-level 4D Reconstruction}
We evaluate scene-level 4D reconstruction on the Kubric dataset~\cite{greff2021kubric} under two settings: (i) multi-object foreground with sparse partial observations, and (ii) full scenes including static background geometry with denser inputs. Input point clouds are downsampled to $N_{\text{in}} = 8192$ to match baseline VAE training settings. As shown in \Cref{tab:kubric_depth_8192}, 
DynaTok achieves the best or tied-best accuracy and substantially improves completeness and normal consistency.
Object-centric 4D methods such as Motion2VecSets~\cite{cao2024motion2vecsets} are not applicable to full scenes due to fixed input size and single-object assumptions. Compared to frame-wise 3D latent baselines, DynaTok achieves substantially better reconstruction quality, highlighting the importance of explicit temporal aggregation.

Qualitative results in \Cref{fig:kubric_scene} further confirm these findings. Frame-wise 3D generative baselines (e.g., TripoSG~\cite{li2025triposg}, 3DShape2VecSet~\cite{zhang20233dshape2vecset}) produce fragmented and noisy geometry due to the lack of temporal integration, whereas DynaTok aggregates partial observations into a coherent canonical space, yielding dense and accurate reconstructions of both dynamic objects and background structures.

\subsection{Ablation Study}
\begin{table}[htbp]
\centering
\caption{\textbf{Ablation Study of \oursName Components on DT4D-A Dataset.}}
\label{tab:ablation_results}
\resizebox{\columnwidth}{!}{%
\begin{tabular}{llccc}
\toprule
\textbf{Component} & \textbf{Configuration} & \textbf{Acc} $\downarrow$ & \textbf{Comp} $\downarrow$ & \textbf{NC} $\uparrow$ \\
\midrule
\multirow{2}{*}{\textbf{Encoder}} 
& Per-frame Only & 0.276 & 0.301 & 0.531 \\
& Geometry-Motion & \textbf{0.027} & \textbf{0.026} &  \textbf{0.877 }\\
\midrule
\multirow{3}{*}{\textbf{\# Tokens}}
& $M=128$ & 0.030 & 0.032 & 0.869 \\
& $M=256$ & 0.027 & \textbf{0.026}  &  0.877 \\
& $M=512$ & \textbf{0.026} & 0.030 & \textbf{0.881} \\
\midrule
\multirow{3}{*}{\textbf{\# Train Views}}
& 2 Views  & 0.031 & 0.035 & 0.864 \\
& 4 Views  & 0.028 & 0.032 & 0.872 \\
& 8 Views  & \textbf{0.027} & \textbf{0.026} & \textbf{0.877} \\
\bottomrule
\end{tabular}
}
\end{table}

We conduct comprehensive ablation studies on the DT4D-A dataset to critically assess the contribution of the architectural innovations within the DynaTok framework, focusing on decoupling effectiveness, tokenization efficiency, and robustness to partial input.

\paragraph{Per-frame (3D) vs Joint (4D) Encoder.}
The central finding confirms the importance of global attention. When the sequence is processed by the ``Per-frame Only'' baseline, performance severely degrades, demonstrating that simple token-wise extraction fails to establish the canonical space and temporal consistency required for decoding. The ``Per-frame Only'' setting trains the model to perform single-frame completion. The substantial gain achieved by the full DynaTok model in~\Cref{tab:ablation_results} ($\mathbf{0.276} \to \mathbf{0.027}$ in Acc) verifies that joint spatiotemporal alignment in the token space is essential for stabilizing geometry and constructing a robust canonical representation for 4D reconstruction.

\paragraph{Number of Latent Tokens ($M$).}
This experiment investigates the critical trade-off between the token count per frame ($M$) and the resulting geometric fidelity and computational efficiency. We varied $M$ across several settings, observing that scaling the token count from a coarse $M=128$ yields substantial gains in precision and coverage. 

\paragraph{Impact of Number of Training Views.}
We further test the generalization capacity of DynaTok by evaluating how input sparsity during training affects the quality of synthesized output. We vary the number of available training views among 2, 4, and 8 views, while testing against a consistent dense 16-view sequence. The quantitative results in ~\Cref{tab:ablation_results} demonstrate a clear dependency: metrics show consistent improvement as the training view count increases. This confirms that exposing the network to richer input views is essential for distilling a highly accurate and robust global geometry token.

\paragraph{Canonical Space Evolution Across Time Steps.}

\begin{figure}[t]
\centering
\includegraphics[width=\linewidth]{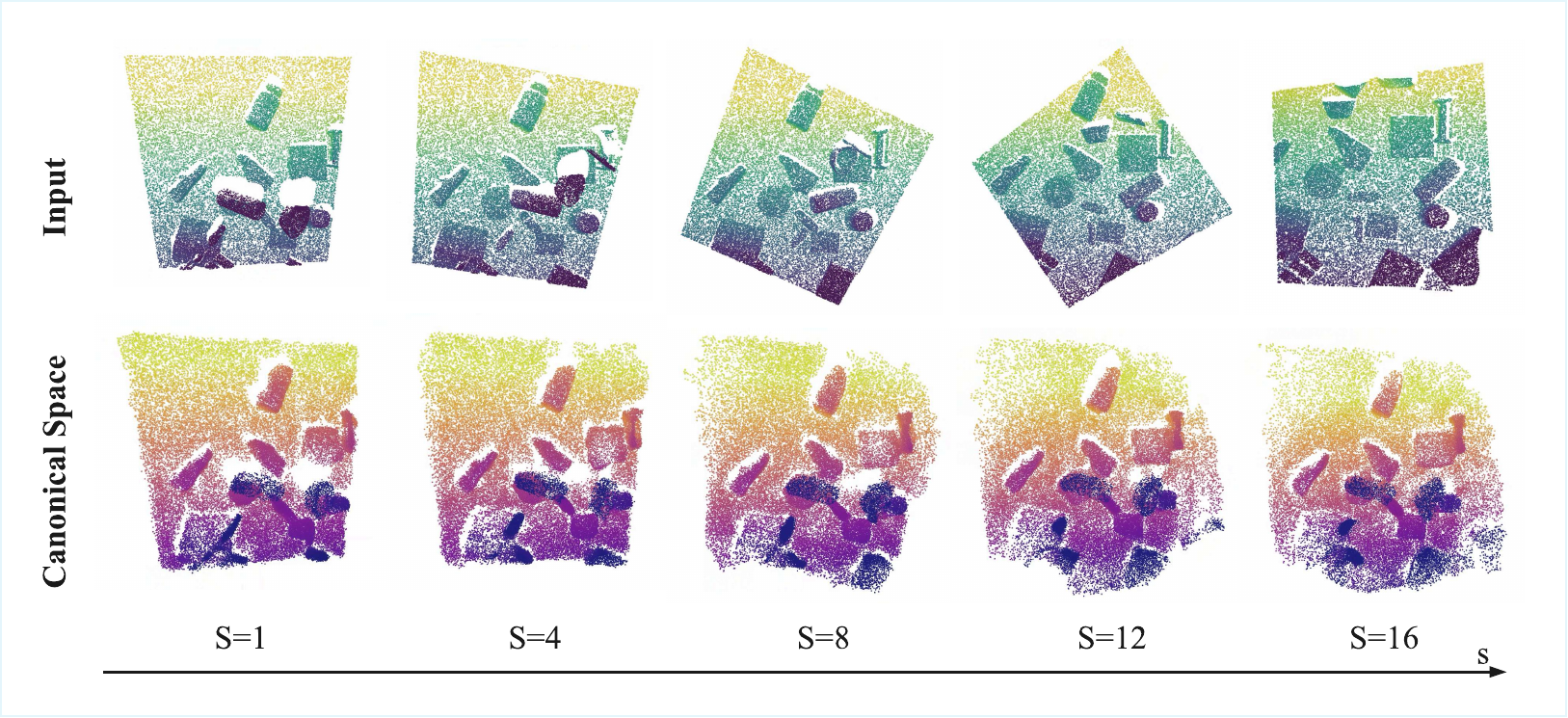}
\caption{\textbf{Qualitative results for canonical space evolution across different time steps.} As additional frames are progressively observed, the canonical space anchored at the reference frame ($s=1$) is expanded consistently to incorporate new geometric information.}%
\label{fig:canonical}
\vspace{-0.8em}
\end{figure}

To analyze temporal aggregation, we study how the canonical representation (first frame) evolves as additional frames are incorporated (S = 1, 4, 8, 12, 16) in~\Cref{fig:canonical}. With only a single frame, the canonical space captures coarse geometry with limited coverage due to sparse observations. As more frames are added, the model progressively aggregates newly observed structures, refining both global shape and local details. The reconstruction becomes increasingly dense, smooth, and complete, indicating effective accumulation of information across time. This behavior confirms that our method fuses multi-frame observations into a unified canonical representation, rather than reconstructing each frame independently.

\paragraph{Model Complexity.}
We further compare model complexity in terms of parameter count, FLOPs, and
inference speed. As shown in Table~\ref{tab:model_complexity}, DynaTok uses
only 19M parameters and requires approximately 6.6k GFLOPs per 16-frame scene.
Inference speed is measured on an L40S GPU using 16-frame sequences. Compared
with Motion2VecSets and TRELLIS, DynaTok is substantially more compact
and computationally efficient, while maintaining strong reconstruction quality.

\begin{table}[t]
\centering
\caption{\textbf{Model complexity comparison}. FLOPs and inference time are measured
for 16-frame scenes, with runtime evaluated on an L40S GPU. DynaTok is more compact and efficient than prior methods.}
\label{tab:model_complexity}
\resizebox{\columnwidth}{!}{
\begin{tabular}{l c c c}
\toprule
Method & Params & GFLOPs & Time / scene \\
\midrule
TRELLIS~\cite{xiang2025trellis} & 1.6B & $\sim$3.7M & $\sim$2 min \\
Motion2VecSets~\cite{cao2024motion2vecsets} & 423M & $\sim$81.7k & $\sim$45s \\
DynaTok (Ours) & 19M & $\sim$6.6k & $\sim$22s \\
\bottomrule
\end{tabular}
}
\end{table}

\subsection{Zero-Shot In-the-Wild Results}
\begin{figure}[t]
\centering
\includegraphics[width=\linewidth]{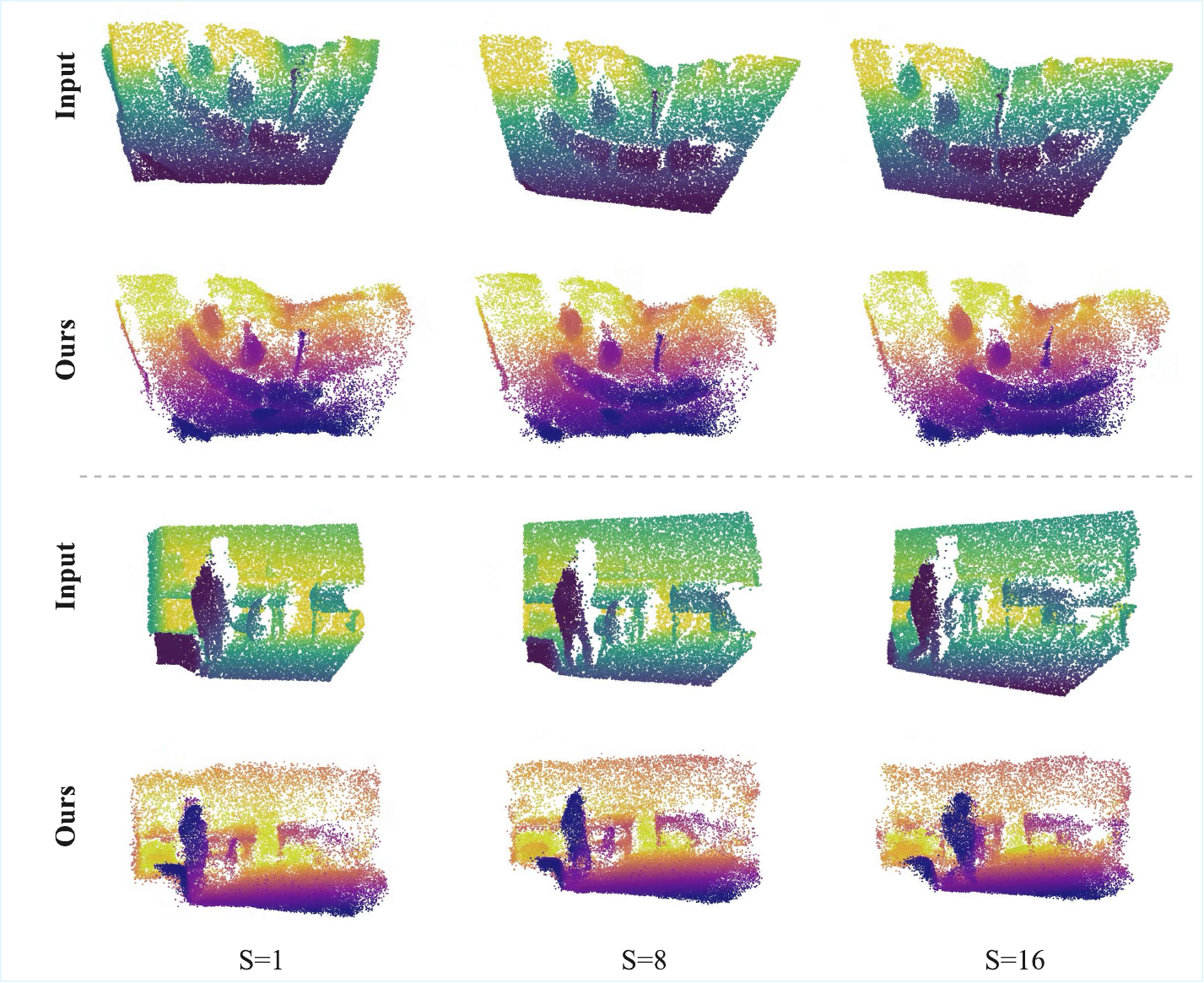}
\caption{\textbf{Qualitative results on in-the-wild data from Bonn RGB-D~\cite{palazzolo2019refusion} and DAVIS~\cite{perazzi2016benchmark}. } We evaluate the generalization ability of our method on real-world scenes using CUT3R’s predictions as input.}%
\label{fig:in_the_wild}
\vspace{-0.8em}
\end{figure}

To evaluate real-world generalization under more challenging input conditions, we test our model without fine-tuning on external datasets, including Bonn RGB-D~\cite{palazzolo2019refusion} and DAVIS~\cite{perazzi2016benchmark}. Instead of using depth sensor measurements, we directly use point maps predicted by the off-the-shelf CUT3R model~\cite{wang2025continuous}, which are noisy, incomplete, and often affected by scale ambiguity. Despite this substantial domain gap from the clean synthetic training data, our method preserves reasonable reconstruction quality and temporal coherence (\Cref{fig:in_the_wild}), demonstrating robustness to noisy and partial real-world observations. 
This suggests a downstream application of DynaTok: 4D fusion with off-the-shelf monocular reconstruction models~\cite{zhang2025efficiently, karhade2025any4d}.

\section{Conclusion}
\label{sec:conclusion}

We introduced DynaTok, a method for 4D reconstruction from partial, correspondence-free point cloud sequences. By emphasizing temporal aggregation in a latent representation, our approach integrates incomplete observations across time to recover coherent 4D geometry without relying on images or explicit tracking. Experiments on object- and scene-level benchmarks demonstrate improved reconstruction quality and temporal consistency under sparse inputs. We believe this work highlights the importance of structured temporal aggregation for point-based 4D reconstruction and provides a foundation for extending dynamic scene modeling to richer sensing modalities such as video and longer temporal horizons.

\section*{Acknowledgements}

We sincerely thank Haofei Xu, Felix Wimbauer, Nando Metzger, Rundi Wu, Lu Sang and Zhaochong An for their insightful feedback, helpful technical discussions, and valuable suggestions during the development of this work.

\section*{Impact Statement}


This paper presents work whose goal is to advance the field of Machine
Learning. There are many potential societal consequences of our work, none of which we feel must be specifically highlighted here.



\bibliography{example_paper}
\bibliographystyle{icml2026}

\newpage
\appendix
\onecolumn

\section{Appendix}
\label{sec:appendix}

\renewcommand{\thetable}{S\arabic{table}}
\setcounter{table}{0}

\subsection{Implementation Details}

\paragraph{Point Token Extractor.}
We adopt the same encoder backbone as the TripoSG VAE encoder~\cite{li2025triposg} to extract latent point tokens. The encoder takes as input a point cloud $\mathbf{X}_s \in \mathbb{R}^{N \times 3}$ and outputs 512 latent tokens of dimension 64. We use one cross-attention layer to fuse spatial features, followed by six self-attention layers to enhance token-wise interactions. During evaluation, we use 8,192 input points per frame, subsampled from valid depth pixels at $224 \times 224$ resolution.

\paragraph{Alignment Module.}
To aggregate information across frames, we use an alignment module composed of eight alternating frame-level and global-level attention blocks. Frame-level attention updates tokens using intra-frame relationships, while global-level attention enables cross-frame aggregation to progressively form geometry and motion tokens. The module outputs 512 geometry tokens and $512 \times S$ motion tokens, each with dimension 64.

\paragraph{Flow-Matching Decoder.}
The decoder is a lightweight 3-layer transformer with self- and cross-attention, similar to~\citep{chang20243d}, and is conditioned on the latent tokens $\mathbf{Z}_s$. It takes an interpolated point cloud $\mathbf{x}_t$ and the flow time $t$ as input, and predicts the conditional velocity field. The decoder uses a hidden dimension of 64. During inference, the final reconstructed 3D point positions are obtained by integrating the learned ODE.

\paragraph{Training and Inference.}
During training, we use 8-frame sequences with 30,000 input points per frame, a batch size of 4 per GPU, a learning rate of $10^{-3}$, and train for 250k iterations. For flow matching, we sample noise from a uniform cube distribution and uniformly sample flow times from $[0,1]$. Following the main text, we define a linear interpolation path from noise to data and train the model to predict the corresponding target velocity field. The interpolated points are Fourier-embedded and decoded by the transformer conditioned on the latent tokens $\mathbf{Z}_s$.

During inference, we solve the learned ODE using 25 Euler steps to generate 50,000 output points. For Kubric evaluation, we use 16-frame sequences and compare the reconstructed point clouds against 100,000 ground-truth target points.

\subsection{Additional Analysis on RGB Input with DINOv2 Tokens}
\label{sec:appendix_rgb_dinov2}

Our work focuses on partial point cloud input, a practical setting motivated by
depth cameras, LiDAR, and 3D scanners. Nevertheless, our framework is modular
and can naturally incorporate RGB cues. To analyze this extension, we use a
frozen DINOv2 encoder to extract image tokens, project them to the point-token
feature dimension, and concatenate them with point-cloud tokens before the
alignment module. The remaining architecture is unchanged. For efficiency, we
train this variant for 10 epochs on Kubric Scene while freezing the point cloud
encoder, DINOv2 encoder, and decoder, and fine-tuning only the projection layer
and alignment module.

Table~\ref{tab:dinov2_rgb_analysis} summarizes the results. RGB-only
conditioning with frozen DINOv2 features already produces reasonable 4D
reconstructions, although it underperforms partial point-cloud input, as DINOv2
features are semantic and not optimized for 3D geometry. When combined with
point clouds, DINOv2 tokens are most useful in the sparse-input regime:
with 1024 input points, they improve the accuracy from 0.014 to 0.013,
Completeness from 0.020 to 0.017, and Normal consistency from 0.855 to 0.866.
With semi-dense input, i.e., 8192 points, the gains are limited and Normal
consistency slightly decreases. These results suggest that RGB cues provide
complementary information mainly when geometric observations are sparse, and
further demonstrate that our framework can be extended to multi-modal inputs in
a shared latent token space with minimal architectural changes.

\begin{table}[t]
\footnotesize
\centering
\caption{Additional analysis on RGB input with DINOv2 tokens. We evaluate both
RGB-only conditioning and RGB-augmented point-cloud input. Frozen DINOv2
features can support reasonable 4D reconstruction from RGB alone and provide
complementary cues under sparse point-cloud input.}
\label{tab:dinov2_rgb_analysis}
\begin{tabular}{l c l c c c}
\toprule
Setting & \# Points & Method & Accuracy $\downarrow$ & Completeness $\downarrow$ & Normal $\uparrow$ \\
\midrule
RGB-only
& -- 
& DINOv2 tokens
& 0.024 & 0.021 & 0.800 \\

Geometry-only
& 8192
& Partial Point Cloud
& 0.010 & 0.014 & 0.880 \\

\midrule
Sparse input
& 1024
& Point-based
& 0.014 & 0.020 & 0.855 \\

Sparse input
& 1024
& Point-based + DINOv2 tokens
& 0.013 & 0.017 & 0.866 \\

Semi-dense input
& 8192
& Point-based
& 0.010 & 0.014 & 0.888 \\

Semi-dense input
& 8192
& Point-based + DINOv2 tokens
& 0.010 & 0.013 & 0.881 \\
\bottomrule
\end{tabular}
\end{table}

\subsection{Further analysis of decoupling and temporal consistency}

We provide two additional quantitative analyses to directly validate
(i) geometry--motion decomposition and (ii) temporal consistency.

\paragraph{Geometry--motion decomposition.}
We directly validate geometry--motion decoupling by comparing decoding with
full tokens, i.e., geometry and motion tokens, against decoding with
geometry-only tokens, where frame-0 tokens are replicated across time.
As shown in Table~\ref{tab:geom_motion_decomp}, the full model remains
stable over time, while geometry-only decoding degrades monotonically.
Moreover, the motion contribution, measured as the Chamfer distance between
geometry-only and full reconstructions, increases steadily over time.
The small nonzero gap at $t=0$ is due to the stochasticity of the
flow-matching decoder. Overall, these results support that geometry tokens
capture the canonical structure, while residual tokens encode the
time-varying deformation required to maintain reconstruction quality.

\begin{table}[t]
\footnotesize
\centering
\caption{Geometry--motion decomposition analysis. We compare full-token
decoding with geometry-only decoding, where frame-0 tokens are replicated
across time. Lower accuracy error is better, while a higher motion contribution
indicates a larger deformation component captured by motion tokens.}
\label{tab:geom_motion_decomp}
\begin{tabular}{c c c c}
\toprule
Frame & Geo-Only Acc. $\downarrow$ & Motion Contribution $\uparrow$ & Full Model Acc. $\downarrow$ \\
\midrule
0  & 0.137 & 0.021 & 0.139 \\
4  & 0.151 & 0.047 & 0.138 \\
8  & 0.162 & 0.071 & 0.137 \\
12 & 0.168 & 0.081 & 0.140 \\
\bottomrule
\end{tabular}
\end{table}

\paragraph{Temporal consistency.}
We measure frame-to-frame consistency using
$\mathrm{CD}(\mathrm{pred}(t), \mathrm{pred}(t+1))$.
As shown in Table~\ref{tab:temporal_consistency}, compared to TRELLIS~\cite{xiang2025trellis},
DynaTok achieves a $2{\times}$ lower mean Chamfer distance and a
$2.5{\times}$ lower variance, indicating substantially more stable temporal
reconstruction.

\begin{table}[t]
\centering
\footnotesize
\caption{Temporal consistency analysis. We report frame-to-frame Chamfer
distance between consecutive predictions. Lower values indicate more stable
temporal reconstruction.}
\label{tab:temporal_consistency}
\begin{tabular}{l c c}
\toprule
Metric & DynaTok & TRELLIS~\cite{xiang2025trellis} \\
\midrule
Mean CD $\downarrow$ & 0.0044 & 0.0089 \\
Std. $\downarrow$   & 0.0007 & 0.0018 \\
Max CD $\downarrow$ & 0.0056 & 0.0124 \\
\bottomrule
\end{tabular}
\end{table}

\subsection{Robustness to Sparse Input Points}
\begin{figure*}[th]
\centering
\includegraphics[width=0.6\linewidth]{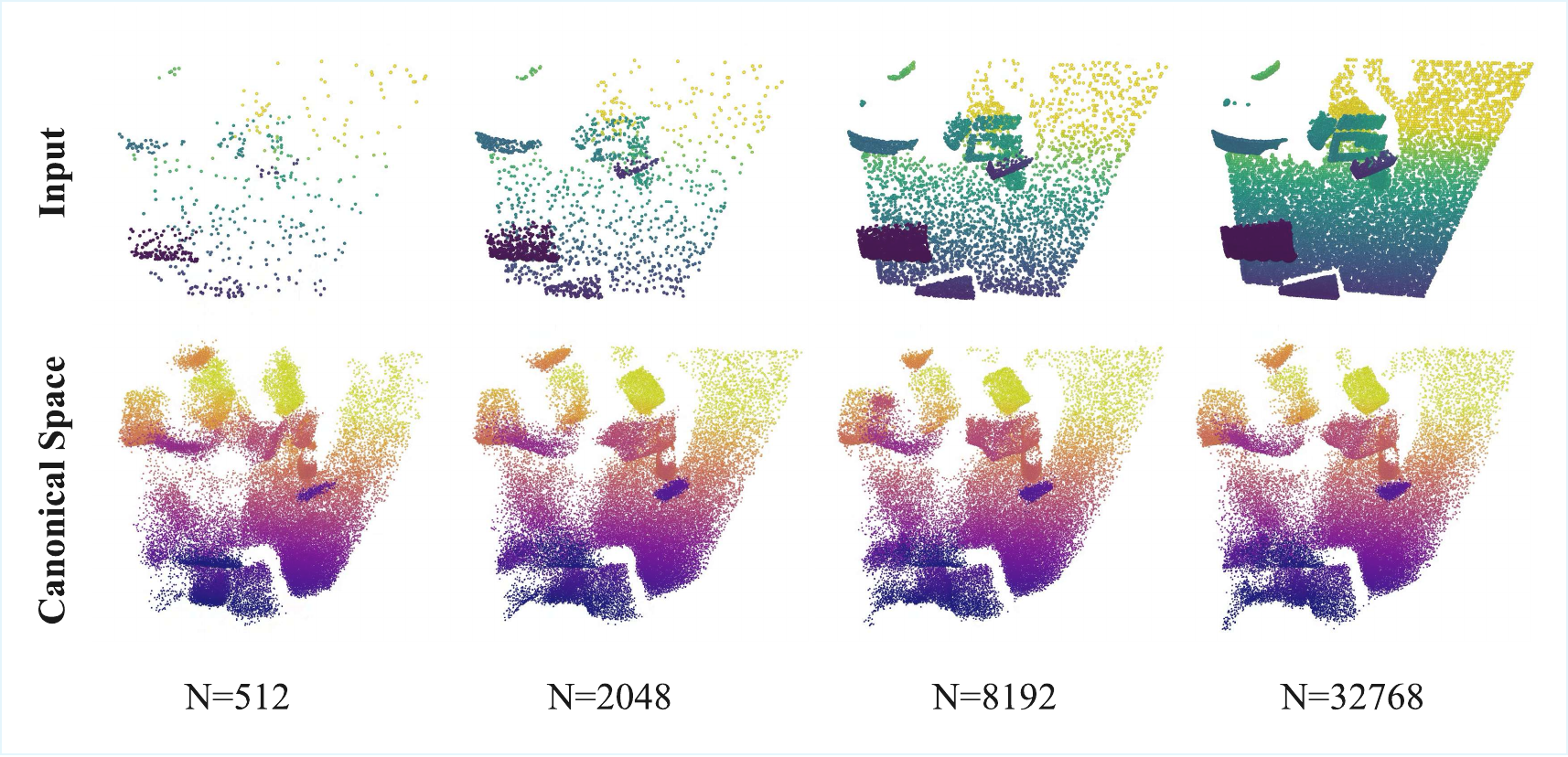}
\caption{\textbf{Qualitative comparison on different numbers of input points.} Our method shows strong robustness to sparse input (e.g., N=512).}%
\label{fig:num_input_pts}
\vspace{-0.8em}
\end{figure*}

To evaluate the robustness of our method under varying input sparsity levels, we conduct experiments using different numbers of input points (N = 512, 2048, 8192, 32768). As shown in~\Cref{fig:num_input_pts}, even when provided with extremely sparse input (e.g., N = 512), our model is still able to recover the overall scene structure and generate a coherent and geometrically plausible reconstruction in the canonical space. As the number of input points increases, the reconstruction gradually becomes denser and more detailed, while maintaining consistent geometry across all settings. This demonstrates that our method does not rely on dense or high-fidelity point inputs and can effectively infer missing structures, highlighting its strong robustness and generalization capability under sparse observations.

\subsection{Discussion of Failure Cases}
Our model is generally robust to short-term occlusions and moderate motion, but it can degrade under prolonged occlusion, very fast motion, heavy object overlap, or motion patterns not represented in the training distribution. The scene model is also less reliable for highly deformable or strongly non-rigid objects, since our Kubric-based training data mainly covers rigid or near-rigid dynamics. Expanding training to larger and more diverse dynamic datasets is an important next step.




\end{document}